\documentclass{article}

\usepackage{microtype}
\usepackage{graphicx}
\usepackage{booktabs} 
\usepackage{caption}
\usepackage{amsmath}
\usepackage{amssymb}
\usepackage{amsfonts}
\usepackage{hyperref}
\usepackage{natbib}
\usepackage{chngcntr}
\usepackage[accepted]{icml2020} 

\begin{document}

\twocolumn[
\icmltitle{A Brief Look at Generalization in Visual Meta-Reinforcement Learning}

\icmlsetsymbol{equal}{*}
\begin{icmlauthorlist}
\icmlauthor{Safa Alver}{mila-mcgill}
\icmlauthor{Doina Precup}{mila-mcgill,deepmind}
\end{icmlauthorlist}

\icmlaffiliation{mila-mcgill}{Mila - McGill University}
\icmlaffiliation{deepmind}{Google DeepMind}
\icmlcorrespondingauthor{Safa Alver}{safa.alver@mail.mcgill.ca}

\icmlkeywords{Machine Learning, Reinforcement Learning}
\vskip 0.3in
]

\printAffiliationsAndNotice{} 

\begin{abstract}
Due to the realization that deep reinforcement learning algorithms trained on high-dimensional tasks can strongly overfit to their training environments, there have been several studies that investigated the generalization performance of these algorithms. However, there has been no similar study that evaluated the generalization performance of algorithms that were specifically designed for generalization, i.e. meta-reinforcement learning algorithms. In this paper, we assess the generalization performance of these algorithms by leveraging high-dimensional, procedurally generated environments. We find that these algorithms can display strong overfitting when they are evaluated on challenging tasks. We also observe that scalability to high-dimensional tasks with sparse rewards remains a significant problem among many of the current meta-reinforcement learning algorithms. With these results, we highlight the need for developing meta-reinforcement learning algorithms that can both generalize and scale.
\end{abstract}

\section{Introduction}

In recent years, deep reinforcement learning (RL) algorithms have achieved significant success in a wide variety of challenging tasks, ranging from board games \cite{silver2017mastering, silver2018general} to video games \cite{mnih2015human, vinyals2017starcraft}. Despite the ever-increasing successes, these algorithms require a substantial amount of data for achieving good performance in a  narrowly-defined domain, and they can perform very poorly even when slight modifications occur in the environment.  This indicates that RL algorithms tend to overfit to the tasks on which they were  trained  \cite{zhang2018study, farebrother2018generalization}. If we want RL algorithms that can learn multiple tasks and quickly adapt to new ones, such algorithms need to learn the common structure across many tasks and then use this information to quickly generalize to new tasks.

Recent studies in the field of meta-reinforcement learning (meta-RL) have shown promising results in this direction \cite{duan2016rl, wang2016learning, finn2017model}. Meta-RL algorithms are trained on multiple related environments, in order to learn a learning algorithm that can perform quick adaptation to new unseen tasks. While these algorithms have shown promise, due to the lack of well-designed benchmarks, they have often been trained and evaluated on very narrow and simple task distributions, leaving their true generalization capabilities unclear. For instance, one of the popular benchmarks involves two tasks which require training a simulated legged robot to run either forward or backward, and another one involves different parametrizations of these robots in terms of their limb configurations \cite{finn2017model, houthooft2018evolved, rakelly2019efficient}. To overcome this problem and study the capabilities of these algorithms, recently \citet{yu2019meta} have designed a simulated robot benchmark with 50 qualitatively-diverse manipulation tasks. However, despite the diversity, the benchmark lacks challenging high-dimensional tasks with sparse rewards, leaving their capabilities in these domains unclear.

In this study, we examine the generalization performance of meta-RL algorithms using challenging vision-based sparse-reward environments. To achieve this, we leverage procedurally generated environments \cite{cobbe2019leveraging}, which allow generating an infinite amount of game levels with greatly varying difficulty. We find that current meta-RL algorithms show strong signs of overfitting when evaluated on challenging environments. We also observe that scalability to high-dimensional tasks remains a significant problem among current meta-RL algorithms, as most of them either perform poorly or run very slowly. Our main contribution is the empirical study of the generalization performance of meta-RL algorithms in vision-based sparse-reward environments. We hope that our findings will stimulate further research progress in improving generalization.

\begin{figure*}
\vskip 0.2in
\begin{center}
\includegraphics[height=6.25cm]{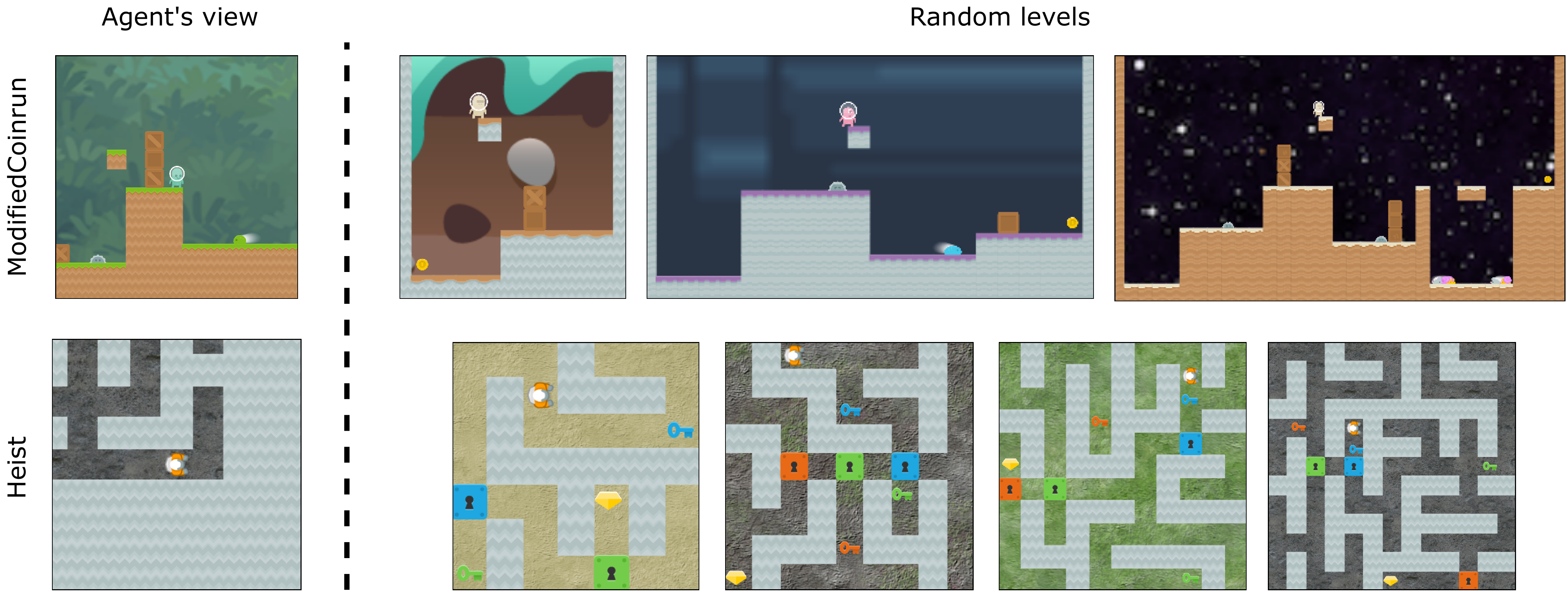}
\caption{(Left) Example observations from the ModifiedCoinrun and Heist environments that the agent receives. Observations consist of only a small patch of space surrounding the agent, making these environments partially observable. (Right) Bird's-eye view of some of the levels from these environments. The difficulty of the levels increases from left to right.}
\label{fig:environments}
\end{center}
\vskip -0.2in
\end{figure*}

The rest of the paper is structured as follows. In section \ref{sec:pre}, we provide a brief preliminary about the formulation we assume in this paper. In section \ref{sec:related}, we review the studies that are most relevant to this study. In section \ref{sec:exp}, we describe our experimental setting and provide our empirical findings. Finally, in section \ref{sec:disc}, we provide a discussion and suggest possible future directions.

\section{Preliminaries}
\label{sec:pre}

\textbf{Reinforcement Learning.} We assume the standard formulation of RL in \citet{sutton2018reinforcement}, where a task is represented by a Markov Decision Process (MDP), a tuple $\mathcal{T} = (\mathcal{S}, \mathcal{A}, p, r, \gamma, T)$. Here $\mathcal{S}$ is the state space, $\mathcal{A}$ is the action space, $p: \mathcal{S} \times \mathcal{A} \times \mathcal{S} \rightarrow \mathbb{R}_{+}$ is the transition distribution, $r: \mathcal{S} \times \mathcal{A} \rightarrow [-R_{\text{max}}, +R_{\text{max}}]$ is the reward function, $\gamma \in (0,1)$ is the discount factor and $T$ is the horizon. The aim of a RL agent is to learn a stochastic policy $\pi_{\theta}: \mathcal{S} \times \mathcal{A} \rightarrow [0,1]$  that  maximizes the expected sum of discounted rewards, which is also known as the return $ \mathbb{E}_{\tau} [ \sum_{k=0}^{T-1} \gamma^k r_{t+k+1} ]$, where $\tau$ denotes trajectories.

\textbf{Meta-Reinforcement Learning.} In meta-RL, we assume a distribution over tasks $p(\mathcal{T})$, where each task is a separate MDP as described above. Importantly, it is often assumed that these tasks share either $\mathcal{S}$ and $\mathcal{A}$ or only $\mathcal{A}$ as in our case. In this setting, the aim of the meta-RL agent is to learn a an update method $U$ from training tasks, which can be learned to generate new stochastic policies $\pi_{U(\theta)}$, allowing quick adaptation to testing tasks. Both training and testing tasks are assumed to be  sampled from $p(\mathcal{T})$.

\section{Related Work}
\label{sec:related}

\textbf{Overfitting in Reinforcement Learning.} Due to the realization that training and testing RL agents in the same environments can prevent the detection of overfitting, there have been a variety of generalization studies where RL agents are evaluated on different environments (mostly different levels and modes of a single game) than those on which they are trained \cite{cobbe2018quantifying, justesen2018illuminating, farebrother2018generalization, zhang2018dissection, zhang2018study, cobbe2019leveraging}. Among these studies, the works of \citet{justesen2018illuminating} and \citet{cobbe2018quantifying, cobbe2019leveraging} are closest to our study. These works use procedural content generation to evaluate generalization in regular RL algorithms. Although our evaluation is inspired by these studies, we look at the generalization problem in the context of meta-RL rather than regular RL.

\textbf{Curriculum Learning in Reinforcement Learning.} The idea of training RL agents using  curricula has been explored widely in the RL literature \cite{schmidhuber2013powerplay, graves2017automated, justesen2018illuminating, matiisen2019teacher, wang2019paired}. In these approaches, the agent is trained with an increasing difficulty of levels. However, in this work we do not consider a structured curriculum, but rather consider one with mixed difficulties (the default setting of Procgen environments \cite{cobbe2019leveraging}). It might also be interesting to explore the  option of increasing task difficulty in the context of meta-RL, but we leave this for future work. Our approach has the advantage of forcing the agent to perform well at all levels of difficulty at the same time.

\begin{figure*}
\vskip 0.2in
\begin{center}
\includegraphics[height=3.75cm]{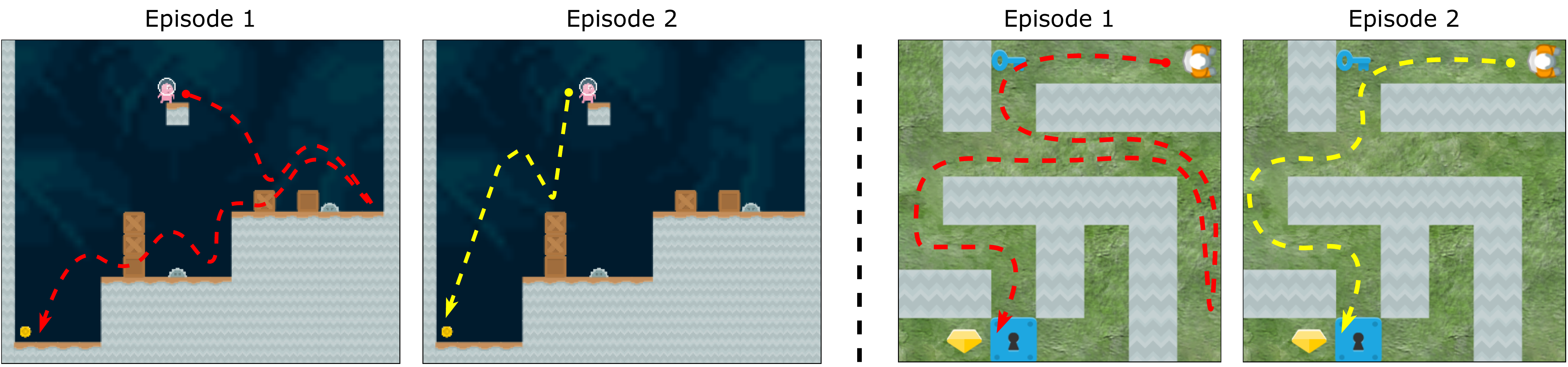}
\caption{Example visualizations of the RL$^2$ agent's behavior in randomly chosen levels. In both of the environments, the agent follows the exploratory red path in the first episode. Then, after discovering where the coin/gem is, in the second episode it follows the yellow path which directly leads to the target. This behavior only arises when given enough training levels.}
\label{fig:path}
\end{center}
\vskip -0.2in
\end{figure*}

\textbf{Meta-Reinforcement Learning Evaluation.} Meta-RL algorithms have been evaluated on a wide range of simulated environments which include 3D maze navigation tasks \cite{duan2016rl, wang2016learning, mishra2017simple}, continuous control tasks with parametric variations \cite{finn2017model, rothfuss2018promp, rakelly2019efficient, houthooft2018evolved, kirsch2019improving}, bandit/MDP problems \cite{duan2016rl, wang2016learning, mishra2017simple}, challenging gridworlds \cite{stadie2018some} and robotic manipulation tasks \cite{yu2019meta}. However, except for the 3D maze navigation tasks, these domains lack two important challenges: high-dimensionality and reward sparsity. Thus, the generalization performance of these algorithms beyond 3D maze navigation tasks remains unclear. In order to investigate this, we evaluate meta-RL algorithms on challenging vision-based, sparse-reward, procedurally generated environments \cite{cobbe2019leveraging}.

\section{Experiments}
\label{sec:exp}

\subsection{Experimental Setting}

\textbf{Environments.} As a testbed, we used two different games from the Procgen environments \cite{cobbe2019leveraging}.\footnote{We have chosen only two environments as the computational cost for our experiments is very high.} Our first choice is a platformer game, Coinrun. However, since the original game requires no exploration, we modified it so that the agent has to perform exploration to find the coin, which is placed either on the far right or far left side of the level. We refer to this modified version as ModifiedCoinrun. Our second choice is a navigation game called Heist, where the agent has to first collect scattered keys to unlock doors, and then reach the gem. In both of these environments, the agent receives a reward of +10 when the coin/gem is picked and the levels end without any reward if the agent dies (only in ModifiedCoinrun) or the 1,000 timesteps limit is reached. Both games are also partially observable, requiring memory (see left side of Fig.~\ref{fig:environments}).

Since meta-RL algorithms are often evaluated in environments where only the level layouts change \cite{duan2016rl, mishra2017simple}, as opposed to the additional changing colors in the Procgen environments, we also created versions of the above environments where only the layout changes across different levels. We refer to these environments as ModifiedCoinrun (Easy Mode) and Heist (Easy Mode). More information on the environments we used can be found in Appendix \ref{app:env}.

An important thing to note is that the levels of the games in Procgen environments are generated deterministically from a given seed, allowing generation of an infinite amount of training and testing levels. They also  greatly vary in difficulty, providing a natural curriculum for learning. Some of the levels, ranging from the easiest to the hardest, are depicted in the right side of Figure \ref{fig:environments}.

\textbf{Algorithm Choice.} For the evaluation process, we experimented with many different meta-RL algorithms that meta-learn a policy. However, we only evaluated RL$^2$ \cite{duan2016rl, wang2016learning} as it was the only algorithm that we were able to successfully train. In our experiments, we found E-RL$^2$ \cite{stadie2018some} to perform very poorly (possibly due to the very delayed reward), SNAIL \cite{mishra2017simple} to require a very large trajectory (which requires an infeasible amount of memory), and PEARL \cite{rakelly2019efficient} to run very slowly. We also do not include MAML \cite{finn2017model}, and the algorithms that build on top of it \cite{stadie2018some, rothfuss2018promp, gupta2018meta}, as \citet{mishra2017simple} has found the computational expense of training MAML in high-dimensional tasks to be prohibitively high. To support  our observations, we would also like to note that except for RL$^2$ and SNAIL, there has been no study that we are aware, at the time of this paper, that reported success of meta-RL algorithms in high-dimensional tasks.

\begin{figure*}
\vskip 0.2in
\centering
\includegraphics[height=4.35cm]{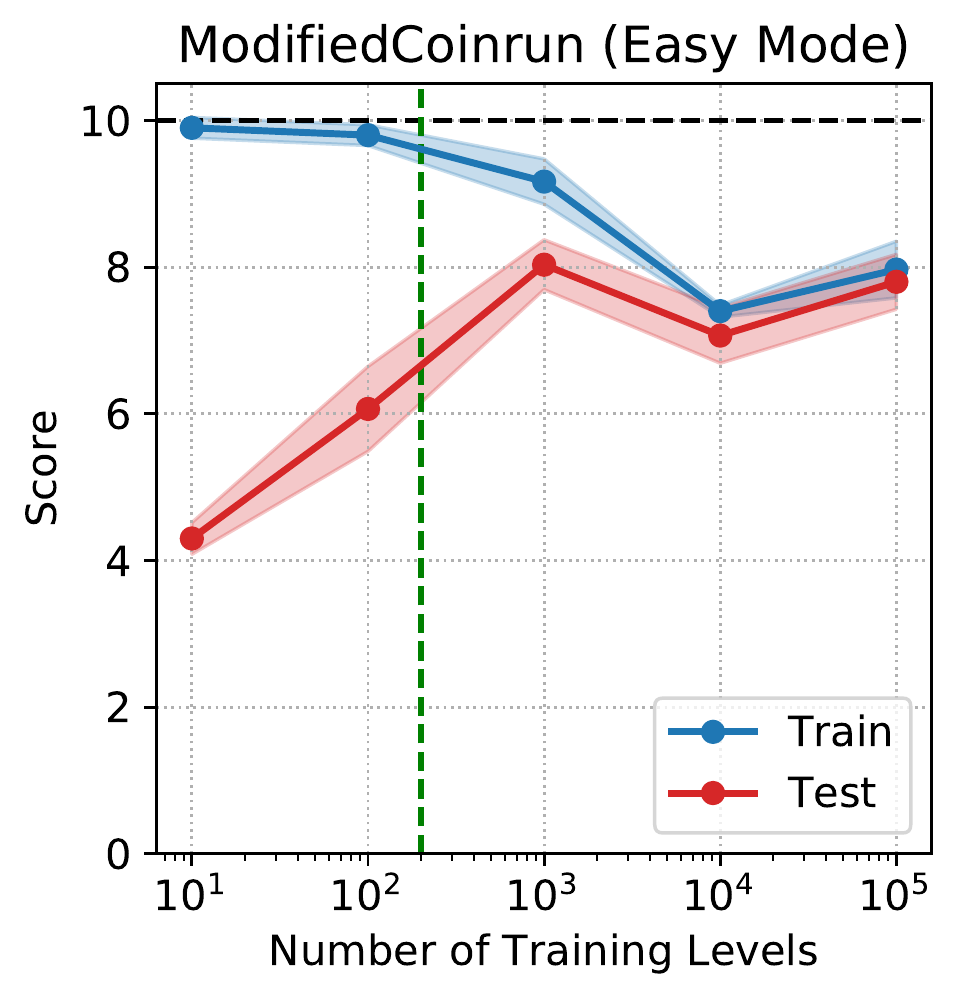}
\includegraphics[height=4.35cm]{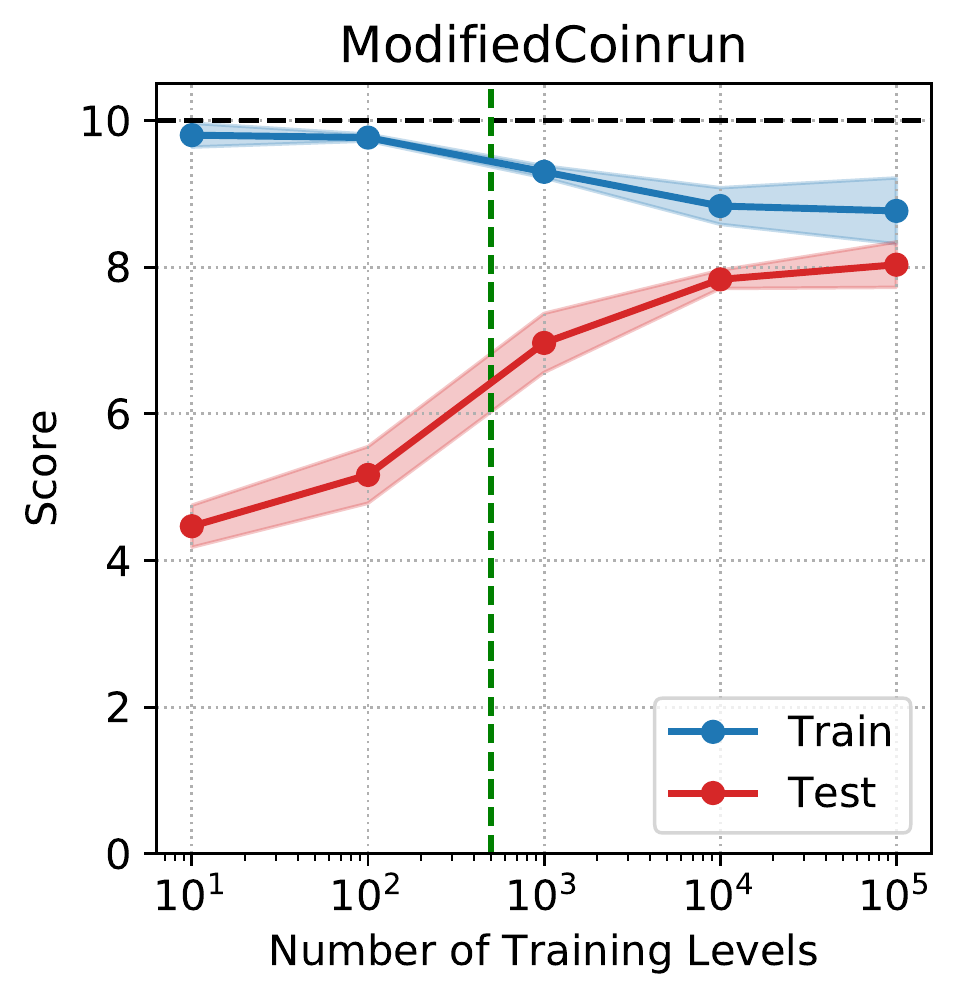}
\includegraphics[height=4.35cm]{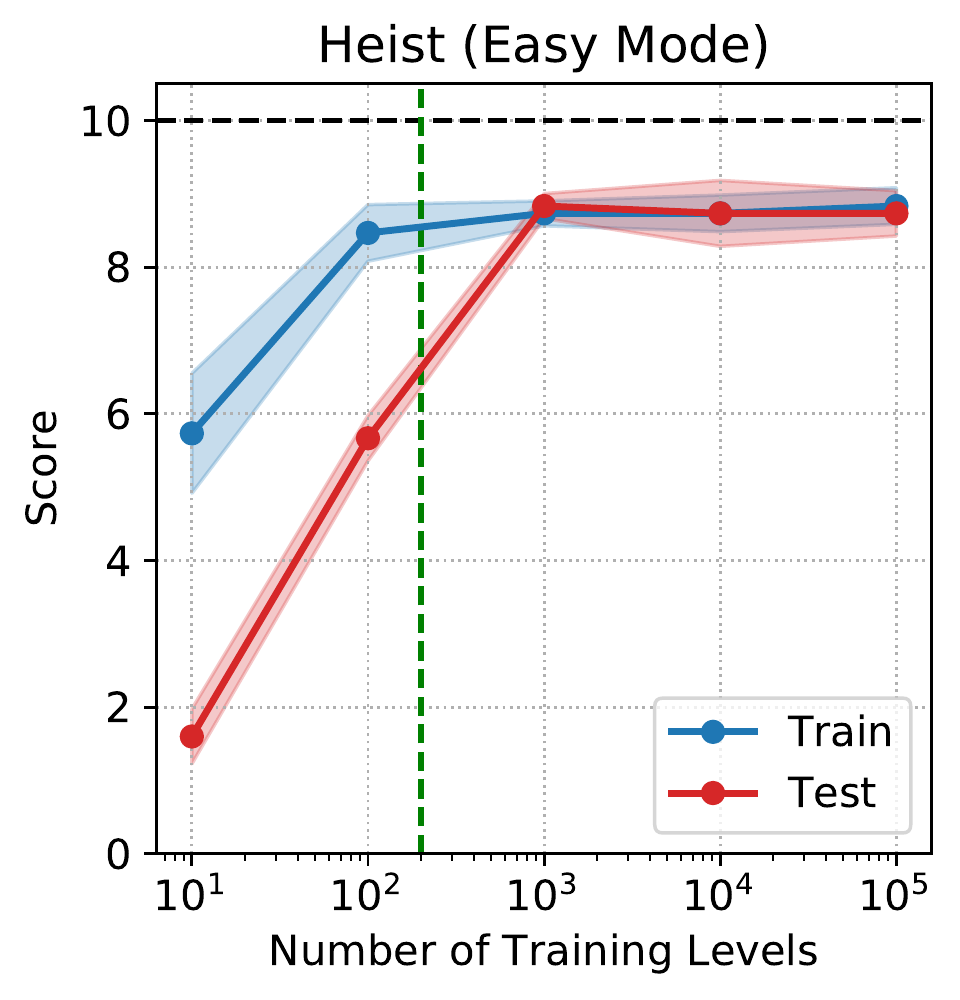}
\includegraphics[height=4.35cm]{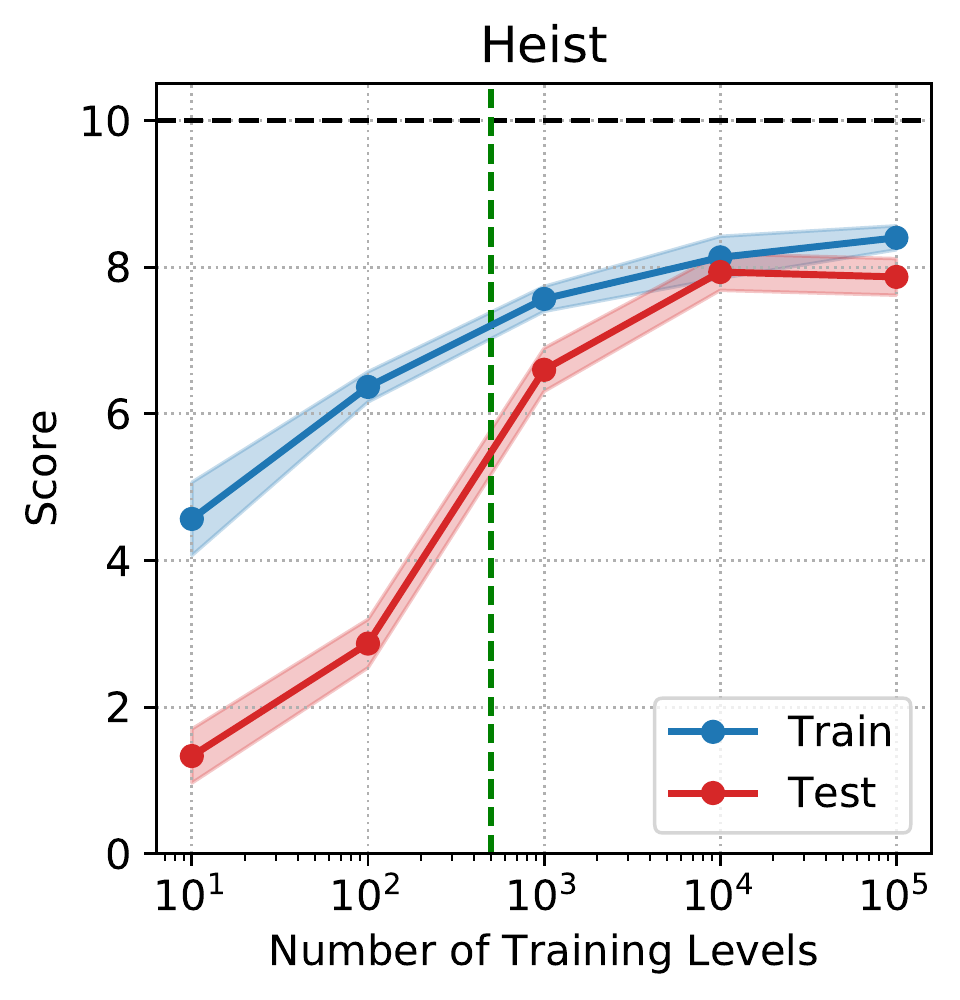}
\includegraphics[height=4.35cm]{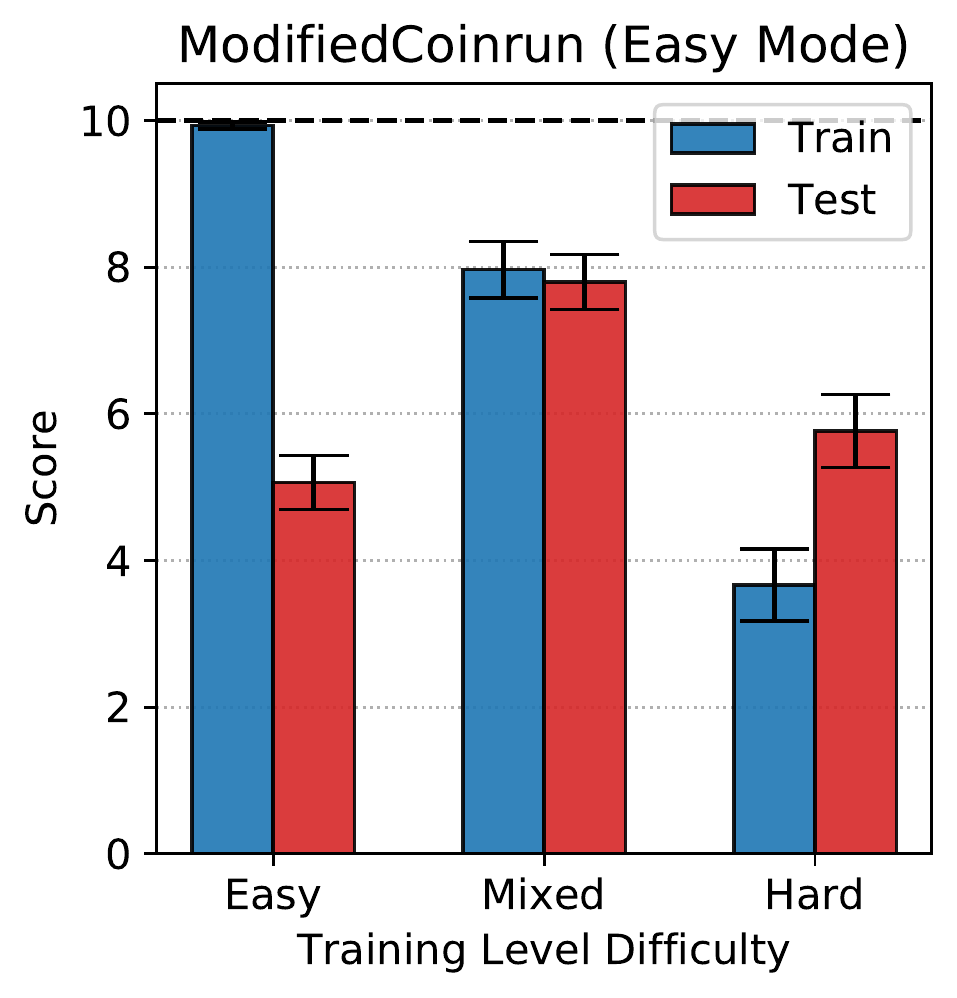}
\includegraphics[height=4.35cm]{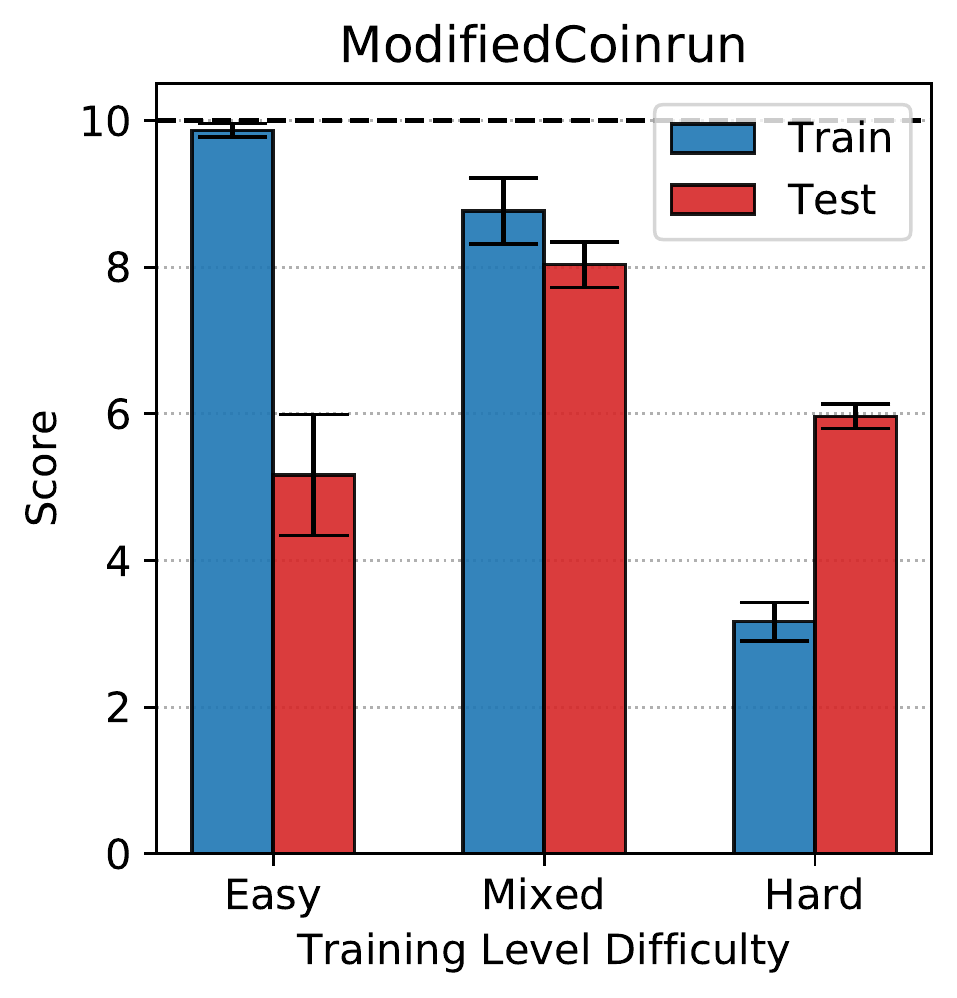}
\includegraphics[height=4.35cm]{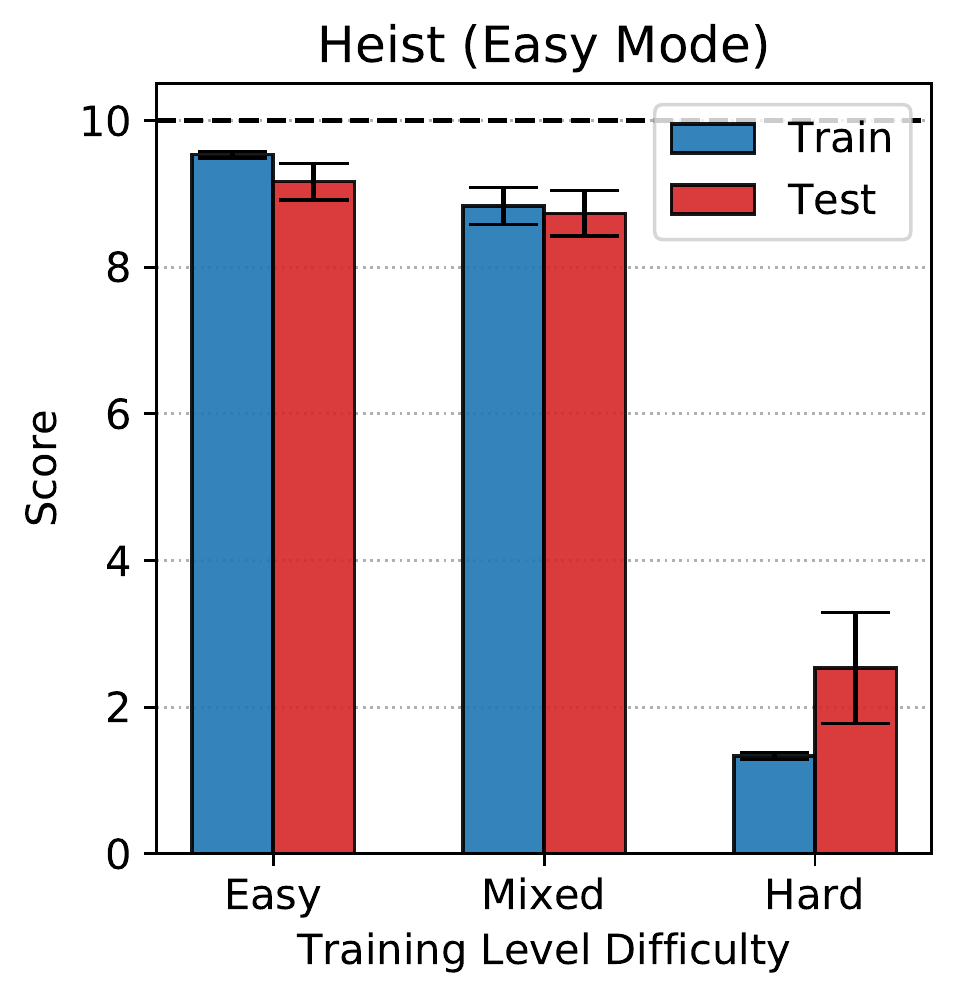}
\includegraphics[height=4.35cm]{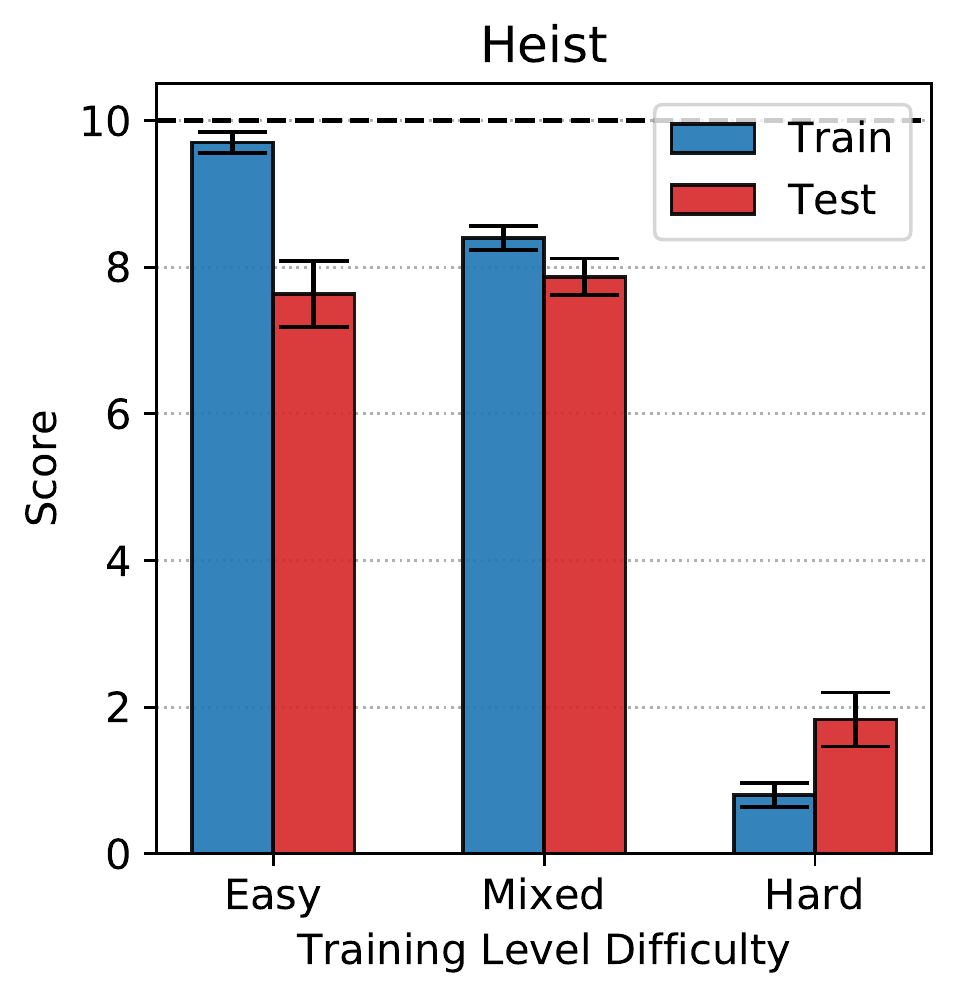}
\caption{(Top row) The final training and test performance of RL$^2$ as a function of the number training tasks. The horizontal black and vertical green dashed lines indicate the maximum achievable scores and recommended training levels, respectively. (Bottom row) The training and test performance for each of the training level difficulties. For each of the plots, we report the mean score in the last episode of the trial. The means and standard deviations are computed using 3 independent runs.}
\label{fig:results}
\vskip -0.1in
\end{figure*}

\textbf{Implementation Details.} Our RL$^2$ implementation, builds on top of the PPO \cite{schulman2017proximal} implementation of \citet{liang2017rllib} and uses the IMPALA CNN \cite{espeholt2018impala} as its visual feature extractor. We use this CNN over the one in \citet{mnih2015human}, as \citet{cobbe2019leveraging} has shown that it performs significantly better on Procgen environments. As in \citet{duan2016rl}, the previous actions, rewards and done signals are fed to the network along with the current observation. We concatenate 2 episodes to form a trial, as we found that adding more episodes does not make any difference in the final performance. In all of our experiments, we trained the RL$^2$ agent for 25M (easy modes) and 100M (regular modes) timesteps. Training more did not increase the performance. More information on the network architecture and hyperparameters can be found in Appendix \ref{app:arch} and \ref{app:hyp}, respectively.

\subsection{Evaluation}

\textbf{Qualitative Pre-evaluation.} Before running generalization experiments, we trained the RL$^2$ agent with 100,000 training levels and tested it on unseen levels to see if it can achieve the optimal behavior, which consists of exploring the level in the first episode, and then, after locating the target, using this information to quickly reach it in the second episode. We indeed observe successful learning of this behavior. Examples of this type of optimal behavior are depicted in Figure \ref{fig:path}.\footnote{Videos of the trained agents on additional levels can be found at \url{https://www.youtube.com/playlist?list=PLcOHCuu_gh7nVISSFKtgT_JZ0iNiIbTqn}.}

\textbf{Generalization Experiment I.} We start our generalization evaluation by looking at the effect of training set size on generalization, as was done in \citet{cobbe2019leveraging} for regular RL algorithms. To investigate this, we constructed sets of training levels, with 10 to 100,000 levels. We then trained RL$^2$ agents on each of these sets for 25M (for easy modes) and 100M (for regular modes) timesteps and tested them on randomly sampled held-out levels. The results are shown in the top row of Figure \ref{fig:results}.

Our results show that small training sets can cause significant overfitting in the meta-RL setting, just as in the regular RL setting. Like regular RL algorithms, meta-RL algorithms also require as many as 10,000 levels to close the generalization gap. More importantly, however, we see that there exists a large generalization gap for the recommended 200 (easy mode) and 500 (regular mode) training levels \cite{cobbe2019leveraging} (see the vertical green dashed lines in the top row of Fig.~\ref{fig:results}). In the Heist environments, this gap is relatively lower only because the training score is also low. This indicates that meta-RL algorithms fail to generalize in the Procgen benchmark, showing strong signs of overfitting instead.

\textbf{Generalization Experiment II.} Next, we investigate generalization on a higher level, by evaluating the effect of having a training set composed of fixed difficulty levels. This time, as the number of levels is not the main concern, we constructed three sets of training levels, each containing 100,000 levels, where the first one contains easy levels, the second one contains mixed difficulty levels and the last one contains hard ones. The details of these sets are available in Appendix \ref{app:env}. We again trained RL$^2$ agents on each of these sets for 25M (for easy mode) and 100M (for regular mode) timesteps and tested them on held out levels with mixed difficulty.

The results are depicted in the bottom row of Figure \ref{fig:results}. We find that while training on hard levels can prevent learning from happening at all, training on easy/mixed levels can have different effects depending on the game. In the Heist environments,  training only on easy levels can allow generalization to unseen mixed levels. This also aligns with the observations of \citet{duan2016rl} that RL$^2$ agents trained on 5$\times$5 mazes can often generalize to 9$\times$9 ones. However, in the ModifiedCoinrun environments, this is not the case, and training on mixed levels is strictly required for generalization. We attribute this finding to the nature of the environments. In the Heist environments, the agent is able to randomly explore in the first episode without dying, whereas in the ModifiedCoinrun environments, random exploration can cause death, with no useful information being passed to the second episode. This suggests that generalization across different level difficulties is possible only in certain environments and strong overfitting at a higher level can occur in certain environments. Investigating more formally what environment characteristics are helpful or detrimental could aid the development of better algorithms.

\section{Discussion and Future Work}
\label{sec:disc}

The results of our experiments show that current meta-RL algorithms can show strong overfitting, despite their explicit goal of generalizing well. We see this behavior even in the simplest settings (easy modes) of the Procgen environments. This matches the recent findings of \citet{yu2019meta}, in which they have found that current meta-RL algorithms can fail to generalize even in the simplest settings of the Meta-World benchmark. If our purpose is to create algorithms that can actually generalize, rather than evaluating them on simple tasks, such as different parametrizations of continuous control tasks or simple maze navigation tasks, we believe that they must be evaluated on challenging environments like Procgen or Meta-World, which can pose significant generalization challenges and can help differentiate different agents. By doing so, we can develop algorithms that can achieve the promise of meta-RL.

Another interesting observation  is that the scalability of current meta-RL algorithms to high dimensional tasks remains a significant problem. Our experiments with a large selection of algorithms have shown that RL$^2$ is the only algorithm that was able to scale to the environments presented in this study. With this, we also highlight the need for developing algorithms with better scaling properties.

In future work, these experiments can be extended to the other games in the Procgen benchmark and possibly to other environments where task creation is easily achieved, resulting in more diverse and challenging benchmarks to evaluate the generalization performance of meta-RL algorithms. Another possible future direction is to investigate the generalization performance of algorithms that are not referred to as meta-RL algorithms, but nevertheless promise generalization.

\bibliography{ms}

\begin{thebibliography}{30}
\providecommand{\natexlab}[1]{#1}
\providecommand{\url}[1]{\texttt{#1}}
\expandafter\ifx\csname urlstyle\endcsname\relax
  \providecommand{\doi}[1]{doi: #1}\else
  \providecommand{\doi}{doi: \begingroup \urlstyle{rm}\Url}\fi

\bibitem[Cobbe et~al.(2018)Cobbe, Klimov, Hesse, Kim, and
  Schulman]{cobbe2018quantifying}
Cobbe, K., Klimov, O., Hesse, C., Kim, T., and Schulman, J.
\newblock Quantifying generalization in reinforcement learning.
\newblock \emph{arXiv preprint arXiv:1812.02341}, 2018.

\bibitem[Cobbe et~al.(2019)Cobbe, Hesse, Hilton, and
  Schulman]{cobbe2019leveraging}
Cobbe, K., Hesse, C., Hilton, J., and Schulman, J.
\newblock Leveraging procedural generation to benchmark reinforcement learning.
\newblock \emph{arXiv preprint arXiv:1912.01588}, 2019.

\bibitem[Duan et~al.(2016)Duan, Schulman, Chen, Bartlett, Sutskever, and
  Abbeel]{duan2016rl}
Duan, Y., Schulman, J., Chen, X., Bartlett, P.~L., Sutskever, I., and Abbeel,
  P.
\newblock Rl$^{2}$: Fast reinforcement learning via slow reinforcement
  learning.
\newblock \emph{arXiv preprint arXiv:1611.02779}, 2016.

\bibitem[Espeholt et~al.(2018)Espeholt, Soyer, Munos, Simonyan, Mnih, Ward,
  Doron, Firoiu, Harley, Dunning, et~al.]{espeholt2018impala}
Espeholt, L., Soyer, H., Munos, R., Simonyan, K., Mnih, V., Ward, T., Doron,
  Y., Firoiu, V., Harley, T., Dunning, I., et~al.
\newblock Impala: Scalable distributed deep-rl with importance weighted
  actor-learner architectures.
\newblock \emph{arXiv preprint arXiv:1802.01561}, 2018.

\bibitem[Farebrother et~al.(2018)Farebrother, Machado, and
  Bowling]{farebrother2018generalization}
Farebrother, J., Machado, M.~C., and Bowling, M.
\newblock Generalization and regularization in dqn.
\newblock \emph{arXiv preprint arXiv:1810.00123}, 2018.

\bibitem[Finn et~al.(2017)Finn, Abbeel, and Levine]{finn2017model}
Finn, C., Abbeel, P., and Levine, S.
\newblock Model-agnostic meta-learning for fast adaptation of deep networks.
\newblock In \emph{Proceedings of the 34th International Conference on Machine
  Learning-Volume 70}, pp.\  1126--1135. JMLR.org, 2017.

\bibitem[Graves et~al.(2017)Graves, Bellemare, Menick, Munos, and
  Kavukcuoglu]{graves2017automated}
Graves, A., Bellemare, M.~G., Menick, J., Munos, R., and Kavukcuoglu, K.
\newblock Automated curriculum learning for neural networks.
\newblock In \emph{Proceedings of the 34th International Conference on Machine
  Learning-Volume 70}, pp.\  1311--1320. JMLR. org, 2017.

\bibitem[Gupta et~al.(2018)Gupta, Mendonca, Liu, Abbeel, and
  Levine]{gupta2018meta}
Gupta, A., Mendonca, R., Liu, Y., Abbeel, P., and Levine, S.
\newblock Meta-reinforcement learning of structured exploration strategies.
\newblock In \emph{Advances in Neural Information Processing Systems}, pp.\
  5302--5311, 2018.

\bibitem[Hochreiter \& Schmidhuber(1997)Hochreiter and
  Schmidhuber]{hochreiter1997long}
Hochreiter, S. and Schmidhuber, J.
\newblock Long short-term memory.
\newblock \emph{Neural computation}, 9\penalty0 (8):\penalty0 1735--1780, 1997.

\bibitem[Houthooft et~al.(2018)Houthooft, Chen, Isola, Stadie, Wolski, Ho, and
  Abbeel]{houthooft2018evolved}
Houthooft, R., Chen, Y., Isola, P., Stadie, B., Wolski, F., Ho, O.~J., and
  Abbeel, P.
\newblock Evolved policy gradients.
\newblock In \emph{Advances in Neural Information Processing Systems}, pp.\
  5400--5409, 2018.

\bibitem[Justesen et~al.(2018)Justesen, Torrado, Bontrager, Khalifa, Togelius,
  and Risi]{justesen2018illuminating}
Justesen, N., Torrado, R.~R., Bontrager, P., Khalifa, A., Togelius, J., and
  Risi, S.
\newblock Illuminating generalization in deep reinforcement learning through
  procedural level generation.
\newblock \emph{arXiv preprint arXiv:1806.10729}, 2018.

\bibitem[Kirsch et~al.(2019)Kirsch, van Steenkiste, and
  Schmidhuber]{kirsch2019improving}
Kirsch, L., van Steenkiste, S., and Schmidhuber, J.
\newblock Improving generalization in meta reinforcement learning using learned
  objectives.
\newblock \emph{arXiv preprint arXiv:1910.04098}, 2019.

\bibitem[Liang et~al.(2017)Liang, Liaw, Moritz, Nishihara, Fox, Goldberg,
  Gonzalez, Jordan, and Stoica]{liang2017rllib}
Liang, E., Liaw, R., Moritz, P., Nishihara, R., Fox, R., Goldberg, K.,
  Gonzalez, J.~E., Jordan, M.~I., and Stoica, I.
\newblock Rllib: Abstractions for distributed reinforcement learning.
\newblock \emph{arXiv preprint arXiv:1712.09381}, 2017.

\bibitem[Matiisen et~al.(2019)Matiisen, Oliver, Cohen, and
  Schulman]{matiisen2019teacher}
Matiisen, T., Oliver, A., Cohen, T., and Schulman, J.
\newblock Teacher-student curriculum learning.
\newblock \emph{IEEE transactions on neural networks and learning systems},
  2019.

\bibitem[Mishra et~al.(2017)Mishra, Rohaninejad, Chen, and
  Abbeel]{mishra2017simple}
Mishra, N., Rohaninejad, M., Chen, X., and Abbeel, P.
\newblock A simple neural attentive meta-learner.
\newblock \emph{arXiv preprint arXiv:1707.03141}, 2017.

\bibitem[Mnih et~al.(2015)Mnih, Kavukcuoglu, Silver, Rusu, Veness, Bellemare,
  Graves, Riedmiller, Fidjeland, Ostrovski, et~al.]{mnih2015human}
Mnih, V., Kavukcuoglu, K., Silver, D., Rusu, A.~A., Veness, J., Bellemare,
  M.~G., Graves, A., Riedmiller, M., Fidjeland, A.~K., Ostrovski, G., et~al.
\newblock Human-level control through deep reinforcement learning.
\newblock \emph{Nature}, 518\penalty0 (7540):\penalty0 529--533, 2015.

\bibitem[Rakelly et~al.(2019)Rakelly, Zhou, Quillen, Finn, and
  Levine]{rakelly2019efficient}
Rakelly, K., Zhou, A., Quillen, D., Finn, C., and Levine, S.
\newblock Efficient off-policy meta-reinforcement learning via probabilistic
  context variables.
\newblock \emph{arXiv preprint arXiv:1903.08254}, 2019.

\bibitem[Rothfuss et~al.(2018)Rothfuss, Lee, Clavera, Asfour, and
  Abbeel]{rothfuss2018promp}
Rothfuss, J., Lee, D., Clavera, I., Asfour, T., and Abbeel, P.
\newblock Promp: Proximal meta-policy search.
\newblock \emph{arXiv preprint arXiv:1810.06784}, 2018.

\bibitem[Schmidhuber(2013)]{schmidhuber2013powerplay}
Schmidhuber, J.
\newblock Powerplay: Training an increasingly general problem solver by
  continually searching for the simplest still unsolvable problem.
\newblock \emph{Frontiers in psychology}, 4:\penalty0 313, 2013.

\bibitem[Schulman et~al.(2017)Schulman, Wolski, Dhariwal, Radford, and
  Klimov]{schulman2017proximal}
Schulman, J., Wolski, F., Dhariwal, P., Radford, A., and Klimov, O.
\newblock Proximal policy optimization algorithms.
\newblock \emph{arXiv preprint arXiv:1707.06347}, 2017.

\bibitem[Silver et~al.(2017)Silver, Schrittwieser, Simonyan, Antonoglou, Huang,
  Guez, Hubert, Baker, Lai, Bolton, et~al.]{silver2017mastering}
Silver, D., Schrittwieser, J., Simonyan, K., Antonoglou, I., Huang, A., Guez,
  A., Hubert, T., Baker, L., Lai, M., Bolton, A., et~al.
\newblock Mastering the game of go without human knowledge.
\newblock \emph{Nature}, 550\penalty0 (7676):\penalty0 354--359, 2017.

\bibitem[Silver et~al.(2018)Silver, Hubert, Schrittwieser, Antonoglou, Lai,
  Guez, Lanctot, Sifre, Kumaran, Graepel, et~al.]{silver2018general}
Silver, D., Hubert, T., Schrittwieser, J., Antonoglou, I., Lai, M., Guez, A.,
  Lanctot, M., Sifre, L., Kumaran, D., Graepel, T., et~al.
\newblock A general reinforcement learning algorithm that masters chess, shogi,
  and go through self-play.
\newblock \emph{Science}, 362\penalty0 (6419):\penalty0 1140--1144, 2018.

\bibitem[Stadie et~al.(2018)Stadie, Yang, Houthooft, Chen, Duan, Wu, Abbeel,
  and Sutskever]{stadie2018some}
Stadie, B.~C., Yang, G., Houthooft, R., Chen, X., Duan, Y., Wu, Y., Abbeel, P.,
  and Sutskever, I.
\newblock Some considerations on learning to explore via meta-reinforcement
  learning.
\newblock \emph{arXiv preprint arXiv:1803.01118}, 2018.

\bibitem[Sutton \& Barto(2018)Sutton and Barto]{sutton2018reinforcement}
Sutton, R.~S. and Barto, A.~G.
\newblock \emph{Reinforcement learning: An introduction}.
\newblock 2018.

\bibitem[Vinyals et~al.(2017)Vinyals, Ewalds, Bartunov, Georgiev, Vezhnevets,
  Yeo, Makhzani, K{\"u}ttler, Agapiou, Schrittwieser,
  et~al.]{vinyals2017starcraft}
Vinyals, O., Ewalds, T., Bartunov, S., Georgiev, P., Vezhnevets, A.~S., Yeo,
  M., Makhzani, A., K{\"u}ttler, H., Agapiou, J., Schrittwieser, J., et~al.
\newblock Starcraft ii: A new challenge for reinforcement learning.
\newblock \emph{arXiv preprint arXiv:1708.04782}, 2017.

\bibitem[Wang et~al.(2016)Wang, Kurth-Nelson, Tirumala, Soyer, Leibo, Munos,
  Blundell, Kumaran, and Botvinick]{wang2016learning}
Wang, J.~X., Kurth-Nelson, Z., Tirumala, D., Soyer, H., Leibo, J.~Z., Munos,
  R., Blundell, C., Kumaran, D., and Botvinick, M.
\newblock Learning to reinforcement learn.
\newblock \emph{arXiv preprint arXiv:1611.05763}, 2016.

\bibitem[Wang et~al.(2019)Wang, Lehman, Clune, and Stanley]{wang2019paired}
Wang, R., Lehman, J., Clune, J., and Stanley, K.~O.
\newblock Paired open-ended trailblazer (poet): Endlessly generating
  increasingly complex and diverse learning environments and their solutions.
\newblock \emph{arXiv preprint arXiv:1901.01753}, 2019.

\bibitem[Yu et~al.(2019)Yu, Quillen, He, Julian, Hausman, Finn, and
  Levine]{yu2019meta}
Yu, T., Quillen, D., He, Z., Julian, R., Hausman, K., Finn, C., and Levine, S.
\newblock Meta-world: A benchmark and evaluation for multi-task and meta
  reinforcement learning.
\newblock \emph{arXiv preprint arXiv:1910.10897}, 2019.

\bibitem[Zhang et~al.(2018{\natexlab{a}})Zhang, Ballas, and
  Pineau]{zhang2018dissection}
Zhang, A., Ballas, N., and Pineau, J.
\newblock A dissection of overfitting and generalization in continuous
  reinforcement learning.
\newblock \emph{arXiv preprint arXiv:1806.07937}, 2018{\natexlab{a}}.

\bibitem[Zhang et~al.(2018{\natexlab{b}})Zhang, Vinyals, Munos, and
  Bengio]{zhang2018study}
Zhang, C., Vinyals, O., Munos, R., and Bengio, S.
\newblock A study on overfitting in deep reinforcement learning.
\newblock \emph{arXiv preprint arXiv:1804.06893}, 2018{\natexlab{b}}.

\end{thebibliography}
\bibliographystyle{icml2020}

\clearpage
\onecolumn
\icmltitle{Supplementary Material}
\appendix
\counterwithin{figure}{section}
\counterwithin{table}{section}

\section{Environment Details}
\label{app:env}

\begin{figure}[h!]
\vskip 0.2in
\begin{center}
\includegraphics[height=4.25cm]{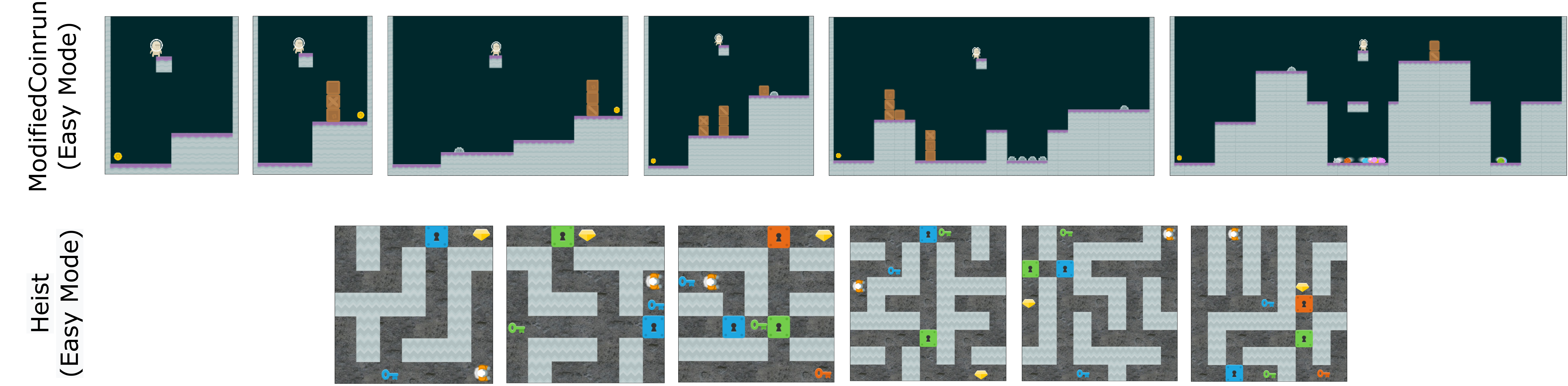}
\caption{Bird's-eye view of some of the levels from the ModifiedCoinrun (Easy Mode) and Heist (Easy Mode) environments. The difficulty increases from left to right. Different from the regular versions of these environments (see Figure \ref{fig:environments}), the background color, agent appearance and wall themes do not change across levels.}
\label{fig:environments_easy}
\end{center}
\end{figure}

\textbf{ModifiedCoinrun.} The ModifiedCoinrun environment is a modified version of the Coinrun environment (hard mode) where the agent spawns on top of a block in the middle of the level, as opposed to always spawning from the far left. This is the only difference from the original environment and it is modified in this way to test the agent's ability to transfer knowledge between episodes. The goal of the agent is again to reach the coin and get a reward of +10. The episode terminates if the agent touches the enemies, lava or saws, or it is not able to get the coin within 1,000 timesteps.

The levels in easy and hard sets are created by setting the \texttt{max\_difficulty} parameter in the source code of Coinrun to 1 (min) and 3 (max), respectively.

\textbf{ModifiedCoinrun (Easy Mode).} The ModifiedCoinrun (Easy Mode) environment is a modified version of the Coinrun environment (easy mode) where again the agent spawns on top a block in the middle of the level. The only difference from the ModifiedCoinrun environment above is that it is created with the flag \texttt{distribution\_mode=easy} rather than \texttt{distribution\_mode=hard}. This flag allows to create different levels where only the level layouts differs between them (see the top row of Fig.~\ref{fig:environments_easy}). 

The levels in easy and hard sets are again created by setting the \texttt{max\_difficulty} parameter in the source code of Coinrun to 1 (min) and 3 (max), respectively.

\textbf{Heist.} The Heist environment is just the original Heist environment with two minor modifications. To make it partially observable and tractable, we create it with the \texttt{distribution\_mode=memory} flag and change the world dimension (controlled by the \texttt{world\_dim} parameter) from 21 to 13, respectively. To goal of the agent is to again pick up the colored keys, unlock the corresponding colored doors and reach the gem to get a reward of +10. The episode terminates after 1,000 timesteps.

The levels in easy and hard sets are created by setting the \texttt{difficulty} parameter in the source code of Heist to 1 (min) and 4 (max), respectively.

\textbf{Heist (Easy Mode).} The Heist (Easy Mode) environment is an easier version of the above Heist environment where the background color is set to a constant image and the world dimension is changed from 13 to 9 (see the bottom row of Fig.~\ref{fig:environments_easy}). 

The levels in easy and hard sets are created by setting the \texttt{difficulty} parameter in the source code of Heist to 1 (min) and 2 (max), respectively.

\section{Network Architecture}
\label{app:arch}

Following \citet{cobbe2019leveraging}, we use the IMPALA CNN architecture \cite{espeholt2018impala} for the visual feature extraction part of our network. We then concatenate the output of the CNN with the vector containing the previous action (one-hoted), previous reward and previous done signal, and pass it through a fully connected layer with 256 units. Finally, we pass this 256 dimensional vector through an LSTM \cite{hochreiter1997long} with 256 units and the output of the LSTM is then fed to two separate fully connected layers (with size 15 and 1) corresponding to the logits for the policy and the value of the state. Except for the last layers, all layers have a ReLU nonlinearity.

\section{Hyperparameters}
\label{app:hyp}

The hyperparameters of the RL$^2$ agent, which is built on top of PPO, is given in Table \ref{tab:hyp}.

\begin{table}[h!]
\caption{Hyperparameters for the RL$^2$ agent.}
\label{tab:hyp}
\centering
\begin{tabular}{l|l} 
 \hline
 Learning rate & $5e^{-4}$ \\ 
 Discount & $0.99$ \\ 
 GAE $\lambda$ & $0.99$ \\ 
 KL coefficient & $0.5$ \\ 
 Target for KL divergence & $0.01$ \\
 Entropy coefficient & $0.01$ \\
 PPO clip parameter & $0.2$ \\
 Gradient clip & $0.5$ \\ 
 Max sequence length & $2000$ \\
 \# of workers & $8$ \\ 
 \# of environments per worker & $16$ \\ 
 \# of episodes in a trial & $2$ \\ 
 \# of SGD iterations & $1$ \\
 Batch size & $16000$ \\ 
 Value function loss coefficient & $0.5$ \\
 \hline
\end{tabular}
\end{table}

\end{document}